\documentclass[]{amsart}
\usepackage{graphicx,amsmath,amssymb,mathtools,booktabs,tabulary,multirow,overpic,xcolor,tikz-cd}
\definecolor{citecolor}{RGB}{34, 139, 34}
\usepackage[pagebackref=true,breaklinks=true,colorlinks,
    citecolor=citecolor,bookmarks=false]{hyperref}
\usepackage[british,american]{babel}
\usepackage[utf8]{inputenc}

\newtheorem{thm}{Theorem}[section]
\theoremstyle{definition}
\newtheorem{definition}[thm]{Definition}
\newtheorem{example}[thm]{Example}

\title[Subspace Match and Learned Representations]{Subspace Match Probably Does Not Accurately Assess the Similarity of Learned Representations}
\author{Jeremiah Johnson}
\address{University of New Hampshire, Manchester, NH 03101}
\email{jeremiah.johnson@unh.edu}
\date{\today}

\begin{document}

\begin{abstract}
Learning informative representations of data is one of the primary goals of deep learning, but there is still little understanding as to what representations a neural network actually learns. To better understand this, subspace match was recently proposed as a method for assessing the similarity of the representations learned by neural networks. It has been shown that two networks with the same architecture trained from different initializations learn representations that at hidden layers show low similarity when assessed with subspace match, even when the output layers show high similarity and the networks largely exhibit similar performance on classification tasks. In this note, we present a simple example motivated by standard results in commutative algebra to illustrate how this can happen, and show that although the subspace match at a hidden layer may be 0, the representations learned may be isomorphic as vector spaces. This leads us to conclude that a subspace match comparison of learned representations may well be uninformative, and it points to the need for better methods of understanding learned representations.
\end{abstract}

\keywords{Neural Networks, Representation Learning, Subspace Match, Algebra, Commutative Algebra}

\maketitle

\section{Introduction}\label{section:introduction}

The intuitive explanation often given for the success of deep learning is that deep neural networks learn `good' representations of data. Indeed, convolutional neural networks are explicitly designed to learn a hierarchical set of representations of images, and aside from only being used for image classification, these hierarchies of representations have been used successfully for image style transfer, object detection, and semantic segmentation, among numerous other tasks \cite{gatys2016image,johnson2017neural,lin2016feature}. However, there is little theory that quantitatively characterizes what neural networks learn.

An early initial attempt at understanding a neural networks learned representation was taken in \cite{li2015convergent}, where representations are studied via correlation analysis and mutual information analysis. In \cite{wang2018towards}, steps are taken toward the development of a rigorous theory for understanding learned representations by considering a simpler problem: given two neural networks with identical architectures trained on the same data in the same way, but with different initializations, how similar are the representations that the networks learn? To assess the similarity of the learned representations, subspace matching is used to characterize neuron activations. Notably, when subspace match is used on neuron activation vectors (see Section \ref{section:preliminaries} for the necessary definitions), we see high similarity at the input and output layers, and low similarity at the hidden layers.

The contribution of this note is to point out via a simple example that the behavior described above should not be unexpected. Furthermore, even where a network has low or no subspace match, it can be the case that the representations are isomorphic. Isomorphism implies structural equivalence from an algebraic perspective; therefore, subspace match likely does not accurately assess the similarity of learned representations. This raises the question, then, of what a better method for assessing similarity of learned representations is, and leaves open the larger question of quantitatively characterizing learned representations.

\section{Preliminaries}\label{section:preliminaries}

Definitions \ref{definition:activation_vector} - \ref{definition:exact_match} follow those given in \cite{wang2018towards}. Let $\mathcal{X}$ and $\mathcal{Y}$ be the set of neurons in the same layer of two networks with identical architecture. Let the input to the networks be $a_1, \dots, a_d$.

\begin{definition}{(Neuron Activation Vector)}\label{definition:activation_vector} Let $v \in \mathcal{X}\cup\mathcal{Y}$. Denote the output of $v$ for input $a_i$ as $z_v(a_i)$. The \emph{activation vector} of $v$ is the vector with $d$ components 
\[
    \mathbf{z}_v = (z_v(a_1), \dots, z_v(a_d)).
\]
We view this vector as the representation of the data learned by neuron $v$ \cite{raghu2017svcca}.
\end{definition}

\begin{definition}{(Representation of a Set of Neurons)}\label{definition:subset_representation} Let $X \subseteq \mathcal{X}$. Let $\mathbf{z}_X := \{\mathbf{z}_x: x \in X\}$. The \emph{representation of} $X$ is 
\[
    \text{span}(\mathbf{z}_x) := \left\{\sum_{\mathbf{z}_x \in \mathbf{z}_X} \lambda_{\mathbf{z}_x}\mathbf{z}_x : \lambda_{\mathbf{z}_x} \in \mathbb{R}\right\} 
\]
The representation of a subset $Y \subseteq \mathcal{Y}$ is defined similarly.
\end{definition}

\begin{definition}{(Exact Match)}\label{definition:exact_match} Let $X \subseteq \mathcal{X}$ and $Y \subseteq \mathcal{Y}$. $X, Y$ is an \emph{exact match} if span$(\mathbf{z}_X) = $span$(\mathbf{z}_Y)$.
\end{definition}

\section{The Algebraic View}\label{section:algebraic_view}

It is a general fact that given a morphism $f:V \to W$, in many cases the morphism factors \cite{lang2002algebra}; that is, there exists $U$ and morphisms $g$ and $h$ such that the diagram below commutes:

\begin{center}
    \begin{tikzcd}
        V \arrow[rd, "f"] \arrow[r, "\exists g"] & U \arrow[d, "\exists h"] \\
                                    & W
    \end{tikzcd}
\end{center}

In fact, we can often go further and find morphisms $\tilde{g} \neq g, \tilde{h} \neq h$ such that the following diagram commutes:

\begin{center}
    \begin{tikzcd}
        V \arrow[rd, "f"] \arrow[r, "g"] \arrow[d, "\tilde{g}"] & U \arrow[d, "h"] \\
        U \arrow[r, "\tilde{h}"]                                & W
    \end{tikzcd}
\end{center}

This implies that the ranges of $g, \tilde{g}$ differ and/or the domains of $h, \tilde{h}$ differ.

A feedforward neural network is a composition of vector space homomorphisms and activation functions, the latter of which are often chosen to be simple nonlinearities such as the ReLU, where ReLU$(\mathbf{x}):= \max{(0, \mathbf{x})}$. Let $W^{(i)}$ denote the weight matrix at the $i^{th}$ layer and $\sigma$ the activation function. Assuming no biases for simplicity, such a network $f$ can be represented as
\begin{equation}\label{equation:composition}
    f(\mathbf{x}) := W^{(k)}(\sigma(W^{(k-1)}(\dots\sigma(W^{(1)}(\mathbf{x}))\dots).
\end{equation}

We can view a feedforward network as a morphism of vector spaces for which we have a factorization provided.

Now denote the post-activation output at layer $i$ of a model with weight matrices $W^{(i)}$ for $i=1,\dots,k-1$ such as that in Equation \ref{equation:composition} as $f_{W^{(i)}}$. That is, 
\[
    f_{W^{(i)}}(\mathbf{x}) := \sigma(W^{(i)}(\sigma(W^{(i-1)}(\dots\sigma(W^{(1)}(\mathbf{x}))\dots).
\]
We wish to choose matrices $W^{(i)}$, $\tilde{W}^{(i)}$ for $i=1,\dots,k$ such that
\begin{align*}
     f(\mathbf{x}) &= W^{(k)}(\sigma(W^{(k-1)}(\dots\sigma(W^{(1)}(\mathbf{x}))\dots) \\
                   &= \tilde{W}^{(k)}(\sigma(\tilde{W}^{(k-1)}(\dots\sigma(\tilde{W}^{(1)}(\mathbf{x}))\dots)
\end{align*}
but $f_{W^{(i)}} \neq f_{\tilde{W}^{(i)}}$ for some $i \in \{1,\dots,k\}$. Doing this will produce an exact match at the set of output neurons, and no match at layer $i$. The following example demonstrates  in a low-dimensional setting that we can do this, effectively producing the $\tilde{g} \neq g$ and $\tilde{h} \neq h$ necessary to complete the commutative diagram above. 

\begin{example}\label{example:the_example} Consider a fully connected feedforward neural network with one hidden layer and two neurons each in the input, hidden, and output layers. Let the weight matrices $W^{(1)}, \tilde{W}^{(1)}, W^{(2)}, \tilde{W}^{(2)}$ be defined as follows:
\begin{equation*}
    W^{(1)}         = \begin{bmatrix*}[r]
                        1 & 0 \\
                        0 & 1
                      \end{bmatrix*}\text{,\hspace{1em}}
    \tilde{W}^{(1)} = \begin{bmatrix*}[r]
                        1 & 0 \\
                        0 & -1
                      \end{bmatrix*}\text{,\hspace{1em}}
    W^{(2)}         = \begin{bmatrix*}[r]
                        1 & -1 \\
                        1 & -1 
                      \end{bmatrix*}\text{,\hspace{1em}}
    \tilde{W}^{(2)} = \begin{bmatrix*}[r]
                        1 & -1 \\
                        1 & -1
                      \end{bmatrix*}.
\end{equation*}

Assume a ReLU activation function, and consider the input data $(1, 1), (-1, -1)$. Then the activation vectors for the neurons at the output layer, $f_{W^{(2)}}$ and $f_{\tilde{W}^{(2)}}$, are identical (the zero vector at each neuron). The activation vectors for the neurons at $f_{W^{(1)}}$ and $f_{\tilde{W}^{(1)}}$ are distinct: for $f_{W^{(1)}}$, the set of activation vectors is $\{[1, 1]^T, [1, 1]^T\}$, while for $f_{\tilde{W}^{(1)}}$ the set is $\{[1, 0]^T, [1, 0]^T\}$. So we have a neural network with identical inputs and identical output activation vectors, but with no subspace match at the hidden layer.

Note that although there is no subspace match at the intermediate layer, the activation vectors at that layer span isomorphic subspaces of $\mathbb{R}^2$: both are lines. The existence of an isomorphism between the two spaces implies that they are structurally similar. This in turn implies that perhaps subspace match is not accurately capturing the similarity of the learned representations.

The weights used in example \ref{example:the_example} were chosen by hand; in current practice, the weights would be determined by some variant of stochastic gradient descent. Given different initializations, there is no reason to suspect that networks with the same architecture could follow paths to different weight matrices and thus different activation vectors in the hidden layers.

\section{Conclusions}\label{section:conclusions}

Example \ref{example:the_example} shows that subspace match is likely not a suitable metric for assessing the similarity of learned representations. There are two reasons why this is the case. First, it is easy to construct examples where the intermediate representations have no match, even when the outputs of the network are identical. Second, even when this is the case, as we can see in example \ref{example:the_example}, the activation vectors at layers with no subspace match may span isomorphic subspaces. The existence of an isomorphism between the subspaces suggests that we should be considering these representations as identical. This brings us back to our original questions: how do we assess the similarity of learned representations, and how can we quantitatively characterize learned representations?

\end{example}

\bibliography{subspace-match}
\bibliographystyle{acm}

\end{document}